\documentclass[preprint,12pt]{elsarticle}

\usepackage{amsmath}
\usepackage[latin1]{inputenc}
\usepackage{multirow}
\usepackage{graphicx}
\usepackage{url}
\usepackage{algorithm}
\usepackage{algorithmic}
\usepackage{rotating}

\begin{document}

\begin{frontmatter}



\title{State-Space Dynamics Distance for Clustering Sequential Data}

\author{Darío García-García\corref{cor1}}
\ead{dggarcia@tsc.uc3m.es}
\ead[url]{http://www.tsc.uc3m.es/~dggarcia}

\author{Emilio Parrado-Hernández}
\ead{emipar@tsc.uc3m.es}

\author{Fernando Diaz-de-Maria}
\ead{fdiaz@tsc.uc3m.es}

\address{Signal Theory and Communications Deparment\\ Escuela Politécnica Superior, Universidad Carlos III de Madrid\\ Avda. de la Universidad, 30\\ 28911, Leganés, Spain}

\cortext[cor1]{Corresponding author. Tel: +34 91 6248805; Fax: +34 91 6248749}


\begin{abstract}
This paper proposes a novel similarity measure for clustering sequential data. We first construct a common state-space by training a single probabilistic model with all the sequences in order to get a unified representation for the dataset. Then, distances are obtained attending to the transition matrices induced by each sequence in that state-space. This approach solves some of the usual overfitting and scalability issues of the existing semi-parametric techniques, that rely on training a model for each sequence. Empirical studies on both synthetic and real-world datasets illustrate the advantages of the proposed similarity measure for clustering sequences.
\end{abstract}


\end{frontmatter}


\section{Introduction}
Clustering is a a core task in machine learning and data processing. It is an unsupervised technique whose goal is to unveil a
natural partition of data into a number of groups or \emph{clusters}. Clustering techniques have been widely studied and
applied for a long time, yielding a large number of well-known and efficient algorithms \cite{xu05}. Recently, the family of algorithms collectively known as spectral clustering \cite{ng02,vonLuxburg07}, that connect the clustering problem with the kernel methods, have received a lot of attention due to their good performance. They share a graph-theoretic approach that results in non-parametric partitions of a dataset, in the sense that they do not require any parametric model of the input data.

The application of clustering techniques to sequential data is lately receiving growing attention \cite{liao05}. In these
scenarios, the useful information is not encoded only 
in the data vectors themselves, but also in the way they evolve along a
certain dimension (usually time). Examples of sequential data range
from stock market analysis to audio signals, video sequences,
etc. Developing machine learning techniques for these scenarios poses
additional difficulties compared to the classical setting where data
vectors are assumed to be independent and identically distributed
(i.i.d.). The common framework consists in combining a  distance
or similarity measure between sequences of variable length, which captures
their dynamical behavior, with a clustering algorithm developed for i.i.d. data.

The design of similarity measures for sequences is generally addressed from a model-based perspective. Specifically,  hidden Markov models (HMMs) \cite{rabiner89} are usually employed as models for the sequences in the dataset. HMMs have been widely used in signal processing and pattern recognition because they oofer a good trade-off between complexity and expressive power. Based on the work of Smyth \cite{smyth97}, many researchers \cite{garcia08,panuccio02,porikli04,yin05} have proposed different distance measures based on a likelihood matrix
obtained using each single sequence to train an HMM. Besides, \cite{jebara07} defines the similarity between two sequences as the probability product kernel (PPK) between the HMMs trained on each sequence. The application of a non-parametric clustering to the distance matrix obtained in one of the aforementioned ways yields a semi-parametric model: each individual sequence follows a parametric model, but no assumption is made about the global cluster structure. These semi-parametric methods have been shown \cite{yin05, jebara07} to outperform both fully parametric methods such as mixture of HMMs \cite{alon03} or combinations of HMMs and
dynamic time warping \cite{oates01}. 

This paper aims to solve a main weakness of the aforementioned semi-parametric
models. The fact that each model is trained using just one sequence can lead
to severe overfitting or non-representative models for short or noisy
sequences. In addition, the learning of a semi-parametric model
involves the  calculation of a number of
likelihoods or probability product kernels that is quadratic in the
number of sequences, which hinders the scalability of the method. To overcome these disadvantages, we propose to train a single
HMM using all the sequences in the dataset, and then cluster the sequences attending to the transition matrices they induce in the state-space of the common HMM. 

The rest of the paper is organized as follows: Section
\ref{sec:clu_seq_hmm} starts with a brief review of hidden Markov
models and spectral clustering followed by a presentation of the state
of the art in model-based clustering of sequences. Section
\ref{sec:algorithm} describes how to cluster sequences using
information about their dynamics in a common state-space. This new proposal is empirically
evaluated in comparison with other methods in Section
\ref{sec:results}. Finally, Section \ref{sec:conclusions} draws the
main conclusions of this work and sketches some promising lines for future research.

\section{Clustering Sequences with Hidden Markov Models}
\label{sec:clu_seq_hmm}
This section reviews the state-of-the-art framework employed to cluster sequential data, which consists of two phases: (1) The design of a similarity or distance measure for sequences based on Hidden Markov Models; and (2) the use of that measure in a clustering algorithm. We opt for
spectral clustering due to its good results reported in the literature. Nonetheless, the distances presented in this work can be used in combination with any clustering algorithm.

\subsection{Hidden Markov Models}
\label{sec:hmms}
Hidden Markov models (HMMs) \cite{rabiner89} are a type of parametric, discrete state-space model.  They provide a convenient model for many real-life phenomena, while allowing for low-complexity algorithms for inference and learning. Their main assumptions are the independence of the observations given the hidden states and that these states follow a Markov chain.

Assume a sequence $\mathbf{S}$ of $T$ observation vectors $\mathbf{S}=\{\mathbf
x_1,\dots,\mathbf x_T\}$. The HMM assumes that $\mathbf{x}_t$, the $t^{th}$ observation
of the sequence, is generated according to the
conditional emission density $p(\mathbf x_t|q_t)$, with $q_t$ being
the hidden state at time $t$. The state $q_t$  can take values from a discrete set $\left\{ s_1, \ldots, s_K \right\}$ of size $K$. The hidden states evolve following a time-homogeneous first-order Markov chain, so that $p(q_t|q_{t-1},q_{t-2},\ldots,q_0)=p(q_t|q_{t-1})$.

In this manner, the parameter set $\theta$ that defines an HMM consists of the following distributions:
\begin{itemize}
\item The initial probabilities vector $\mathbf{\pi}=\left\{ \pi_i \right\}_{i=1}^K$,  where $\pi_i=p(q_0=s_i)$.
\item The state transition probability, encoded in a matrix $\mathbf{A}=\left\{a_{ij}\right\}_{i,j=1}^K$ with $a_{ij}=p(q_{t+1}=s_j|q_t=s_i)$, $1 \leq i,j \leq K$.
\item The emission pdf for each hidden state $p(\mathbf{x}_t|q_t=s_i), 1 \leq i \leq K$.
\end{itemize}

From these definitions, the likelihood of a sequence $\mathbf S = \{\mathbf
x_1,\dots,\mathbf x_T\}$ can be written in the following factorized way:
\begin{equation}
\label{eq:lik_hmm}
p(\mathbf{S}|\theta)=\sum_{q_0,\ldots,q_T}\pi_{q_0}p(\mathbf
x_0|q_0)\prod_{t=1}^T p(\mathbf x_t|q_t)a_{q_{t-1},q_t}.
\end{equation}

The training of this kind of models in a maximum likelihood setting is usually accomplished using the Baum-Welch method \cite{rabiner89}, which is a particularization of the well-known EM algorithm. The E-step finds the expected state occupancy and transition probabilities, which can be done efficiently using the forward-backward algorithm \cite{rabiner89}. This algorithm implies the calculation of both the forward $\alpha$ and backward $\beta$ variables that are defined as follows:
\begin{eqnarray}
\alpha_k(t)&=&p(\mathbf x_1,\ldots,\mathbf x_t,q_t=s_k)
\\
\beta_k(t)&=&p(\mathbf x_{t+1},\ldots,\mathbf x_T|q_t=s_k).
\end{eqnarray}
These variables can be obtained in $O(K^2T)$ time through a recursive procedure and can be used to rewrite the likelihood from  Eq. (\ref{eq:lik_hmm}) in the following manner:
\begin{equation}
p(\mathbf{S}|\theta)=\sum_{k=1}^K \alpha_k(t)\beta_k(t),
\end{equation}
which holds for all  values of $t \in \{1,\dots,T\}$. 

Given a previously estimated $\mathbf{A}$, the state transition probabilities can be updated using the forward/backward variables and that previous estimation, yielding:
\begin{equation}
\label{eq:chi}
\tilde{a}_{ij} \propto
\sum_{t'=1}^T\alpha_i(t')a_{ij}p(\mathbf x_{t'+1}|q_{t'+1}=s_j)\beta_j(t'+1).
\end{equation}
Then, the M-step updates the parameters in order to maximize the likelihood given the expected hidden states sequence. These two steps are then iterated until convergence. It is worth noting that the likelihood function can have many local maxima, and this algorithm does not guarantee convergence to the global optimum. Due to this, it is common practice to repeat the training several times using different initializations and then select as the correct run the one providing a larger likelihood.

The extension of this training procedure to multiple input sequences is straightforward. The interested reader is referred to \cite{rabiner89} for a complete description.


\subsection{Spectral Clustering}
\label{sec:sc}
Clustering \cite{xu05} consists in partitioning a dataset $\mathcal{S}$ comprised of $N$ elements into $C$ disjoint groups called clusters. Data assigned to the same cluster must be similar and, at the same time, distinct from data assigned to the rest of clusters. It is an unsupervised learning problem, meaning that it does not require any prior labeling of the data, and thus it is very appropriate for exploratory data analysis or scenarios where obtaining such a labeling is costly.

Algebraically, a clustering problem can be formulated in the following way. Given a
dataset $\mathcal{S}$, one forms a $N\times N$ similarity matrix $\mathbf{W}$, whose $ij^{th}$ element $w_{ij}$ represents the similarity between the $i^{th}$ and $j^{th}$ instances.  The clustering
problem then consists in obtaining a $N \times C$ clustering matrix $\mathbf{Z}$, where 
$z_{ic}=1$ if instance $i$  belongs to cluster $c$ and $z_{ic}=0$ otherwise, which is optimal under some criterium.

Spectral clustering (SC)  algorithms \cite{vonLuxburg07} approach the clustering task from a graph-theoretic perspective.  Data instances form the nodes  $V$ of a weighted graph $G=(V,E)$ whose edges $E$ represent the similarity or adjacency between data, defined by the matrix $\mathbf{W}$. This way, the clustering problem is cast into a graph partitioning one. The clusters are given by the partition of $G$ in $C$ groups that optimize certain criteria such as the normalized cut \cite{shi00}. Finding such an optimal partition is an NP-hard problem, but it can be relaxed into an eigenvalue problem on the Laplacian matrix $\mathbf{L}=\mathbf{D}-\mathbf{W}$, where $\mathbf{D}$ is the diagonal matrix with elements $d_{ii}=\sum_{j=1}^N w_{ij}$, or one of its normalized versions, followed by
$k$-means or any other clustering algorithm on the rows of the matrix of selected eigenvectors. Specifically, in the experiments in this paper, we employ the spectral clustering algorithm defined in \cite{ng02}, which uses the symmetric version of the normalized Laplacian $\mathbf{L}_{\mbox{\small sym}}=\mathbf{D}^{-1/2}\mathbf{L}\mathbf{D}^{-1/2}$. The actual clustering is applied to the normalized rows of the $C$ eigenvectors associated with the lowest eigenvalues. It is worth noting that the embedding of the original data into the rows of the normalized Laplacian eigenvectors also have some appealing properties from the point of view of dimensionality reduction \cite{belkin01}.

The time complexity for the spectral clustering is dominated by the eigendecomposition of the normalized Laplacian, which in general is $O(N^3)$. However, if the affinity matrix is sparse (e.g. if only the affinities between the nearest neighbors of a given node are considered), there exist efficient iterative methods that notably reduce this complexity, such as the Lanczos method \cite{golub89}, which makes it feasible even for large datasets \cite{vonLuxburg07}.

\subsection{Semi-Parametric Model-Based Sequence Clustering}
\label{sec:previous}
Semi-parametric sequence clustering methods make no assumptions about the cluster structure. These methods typically use generative models such as HMMs in order to represent the sequences in the dataset in a common space, and then apply a non-parametric clustering algorithm. Following \cite{smyth97}, many methods use a log-likelihood matrix $\mathbf{L}$ whose ${ij}^{th}$ element $l_{ij}$ is defined as 
\begin{equation}
\label{eq:L}
l_{ij}=\log{p_{ij}}=\frac{1}{\mbox{length}(\mathbf{S}_j)}\log{p(\mathbf{S}_j|\theta_i)}, \quad 1\leq i,j \leq N,
\end{equation}
where $\theta_i$ is the model trained for the $i^{th}$ sequence
$\mathbf{S}_i$ and $N$ is the number of sequences in the dataset. The log-likelihoods are normalized with respect to the length
of the sequences to compensate the exponentially increasing size of the probability space. 

The literature reports several methods to construct a similarity (distance) matrix $\mathbf{D}$
from $\mathbf L$, such as:
\begin{itemize}
\item The ``symmetrized distance'' (SYM) \cite{smyth97}: $d_{ij}^{\mbox{\tiny SYM}}=\frac{1}{2}(l_{ij}+l_{ji})$.
\item The BP distance  in \cite{panuccio02}: 
$d_{ij}^{\mbox{\tiny
    BP}}=\frac{1}{2}\left\{\frac{l_{ij}-l_{ii}}{l_{ii}}+\frac{l_{ji}-l_{jj}}{l_{jj}}\right\}$,
which takes into account how well a model represents the sequence it has been trained on.
\item The Yin-Yang distance of \cite{yin05} $d_{ij}^{\mbox{\tiny YY}}=\left|l_{ii}+l_{jj}-l_{ij}-l_{ji}\right|$,
\end{itemize}
The distance matrix $\mathbf{D}$ is then fed into a clustering algorithm that
partitions the set of sequences. It is worth noting that all these methods
define the distance between two sequences $\mathbf{S}_i$ and $\mathbf{S}_j$ using solely
the models trained on these particular sequences ($\theta_i$ and $\theta_j$).

In \cite{jebara07} another method for constructing the similarity matrix in a model-based approach is proposed which avoids the calculation of the likelihood matrix $\mathbf{L}$. Again an HMM is trained on each individual sequence, but then similarities between sequences are computed directly through a probability product kernel (PPK). Specifically, this kernel is obtained by integrating a product of the probability distributions spanned by two HMMs in the space of the sequences of a fixed length $T_{\mbox{\tiny PPK}}$, which is a free parameter. This way, $\mathbf{L}$ is no longer necessary because the similarities are obtained directly from parameters of the models. The calculation of the PPK between two HMMs can be carried out in $O(K^2T_{\mbox{\tiny PPK}})$ time. 

Furthermore, the method proposed in \cite{garcia08} assumes the existence of a latent model space $\boldsymbol{\theta}$ formed by some HMMs that actually span the data space. Then, the models $\theta_1,\ldots,\theta_N$ trained on each sequence are regarded as a set $\boldsymbol{\tilde{\theta}}$ of intelligently sampled points of $\boldsymbol{\theta}$. Let $\mathbf{L}_N$ be a column-wise normalisation of  $\mathbf{L}$. The $i^{th}$ column of $\mathbf{L}_N$ can be interpreted as a probability density function $f_{\tilde{\boldsymbol{\theta}}}^{\mathbf{S}_1}(\theta)$ over the approximated model space $\boldsymbol{\tilde{\theta}}$ for $\mathbf{S}_i$. This way it is possible to re-express $\mathbf{L}_N$ as:
\begin{displaymath}
\mathbf{L}_N=\left[f_{\tilde{\boldsymbol{\theta}}}^{\mathbf{S}_1}(\theta),\ldots,f_{\tilde{\boldsymbol{\theta}}}^{\mathbf{S}_N}(\theta)\right].
\end{displaymath} 
This interpretation leads to a distance measure consisting in Kullback-Leibler (KL) divergences between the columns of $\mathbf{L}_N$, so:
\begin{equation}
d_{ij}^{\mbox{\tiny KL}}=D_{\mbox{\tiny KL}_{\mbox{\tiny SYM}}}\left(f_{\tilde{\boldsymbol{\theta}}}^{\mathbf{S}_i}||f_{\tilde{\boldsymbol{\theta}}}^{\mathbf{S}_j}\right),
\end{equation}
where $D_{\mbox{\tiny KL}_{\mbox{\tiny SYM}}}$ stands for the symmetrized version of the KL divergence: $D_{\mbox{\tiny KL}_{\mbox{\tiny SYM}}}(p||q)=\frac{1}{2}\left(D_{\mbox{\tiny KL}}(p||q)+D_{\mbox{\tiny KL}}(q||p)\right)$. This way, the distance between two sequences is obtained in a global way, using a data-dependent representation. Under this paradigm, it is possible to select a subset of $P\leq N$ sequences to train individual models on, instead of fitting an HMM to every sequence in the dataset. This can reduce the computational load and improve the performance in real-world data. However, the a priori selection of such a subset remains an open problem. 

The aforementioned semi-parametric methods share the need to train an individual HMM on each sequence in the dataset (or in a subset of it, as in \cite{garcia08}). We see this as a disadvantage for several reasons. Short sequences are likely to result in overfitted models, giving unrealistic results when used to evaluate likelihoods or PPKs. Moreover, training individual models in an isolated manner prevents the use of similar sequences to get more accurate representations of the states. As for the computational complexity, these methods do not scale well with the dataset size $N$. Specifically, the number of likelihoods that need to be obtained is $N^2$ (or $PN$ using the KL method). In the case of PPKs, $N^2/2$
evaluations are required, since the kernel is symmetric.

\section{State Space Dynamics (SSD) Distance}
\label{sec:algorithm}
In this paper, we propose to take a different approach in order to
overcome the need to fit an HMM to each sequence. To this end, we
propose to train a single, large HMM $\theta$ of $K$ hidden states
using all the sequences in the dataset. This will allow for a better
estimation of the emission probabilities of the hidden states,
compared to the case where an HMM is trained on each sequence. Then,
we use the state-space of $\theta$ as a common representation for the
sequences. Each sequence $\mathbf{S}_n$ is linked to the common state-space through
the transition matrix that it induces when is fed into the model. This matrix is denoted as $\mathbf{\tilde{A}}^n=\left\{a^n_{ij}\right\}_{i,j=1}^K$, where
\begin{equation}
\tilde{a}^n_{ij}=p(q^n_{t+1}=s_j|q^n_t=s_i,\mathbf{S}_n,\theta).
\end{equation}

In order to obtain each $\mathbf{\tilde{A}}^n$, we run the forward-backward algorithm for sequence $\mathbf{S}_n$ under the parameters $\theta$ (including the learned transition matrix $\mathbf{A}=\left\{a_{ij}\right\}$) and then obtain the sequence-specific transition probabilities by using equation (\ref{eq:chi}):

\begin{equation}
\tilde{a}^n_{ij} \propto
\sum_{t'=1}^T\alpha^n_i(t')a_{ij}p(\mathbf x_{t'+1}|q_{t'+1}=s_j)\beta^n_j(t'+1),
\end{equation}
where $\alpha^n_i(t)$ and $\beta^n_j(t'+1)$ are the forward and backward variables for $\mathbf{S}_n$, respectively. This process can be seen as a projection of the dynamical characteristics of $\mathbf{S}_n$ onto the state-space defined by the common model $\theta$. Therefore, the overall transition matrix $\mathbf{A}$ of the large model $\theta$ act as a common, data-dependent ``prior'' for the estimation of these individual transition matrices. 

This procedure is somewhat equivalent to obtaining individual HMMs with emission distributions that are shared or ``clamped'' amongst the different models. Clamping is a usual and useful tool when one wants to reduce the number of free parameters of a model in order to either obtain a better estimate or reduce the computational load. In our case, the effects of clamping the emission distributions are two-fold: we get the usual benefit of better estimated parameters and, at the same time, it allows for simple distance measures between hidden Markov models using the transition distributions. This happens because the transition processes of the different models now share a common support, namely the fixed set of emission distributions. 

As previously mentioned, running the forward-backward algorithm implies a time complexity of $O(K^2T)$ for a sequence of length $T$, which is the same complexity required for obtaining the likelihood of an HMM. Our proposal only requires $N$ of these calculations, instead of $N^2$ likelihood evaluations or $N^2/2$ PPKs as in the methods mentioned in the previous section. This makes the SSD algorithm a valuable method for working with large datasets.

At this point, we have each sequence $\mathbf{S}_n$ represented by its induced transition matrix $\mathbf{\tilde{A}}^n$.
In order to define a meaningful distance measure between these matrices,  we can think of each  $\mathbf{\tilde{A}}^n=\left[\mathbf{a}_{n1},\ldots,\mathbf{a}_{nK}\right]^T$ as a collection of $K$ discrete probability functions $\mathbf{a}_{n1},\ldots,\mathbf{a}_{nK}$, one per row, corresponding with the transition probabilities from each state to every other state. In this manner, the problem of determining the affinity between sequences can finally be transformed into the well-studied problem of measuring similarity between distributions. In this work, we employ the Bhattacharyya affinity \cite{bhat43}, defined as:
\begin{equation}
D_B(p_1,p_2)=\sum_x{\sqrt{p_1(x)p_2(x)}},
\end{equation}
where $p_1$ and $p_2$ are discrete probability distributions. We consider the affinity between two transition matrices to be the mean affinity between their rows. The distance between two sequences $\mathbf{S}_i$ and $\mathbf{S}_j$ can then be written as:
\begin{equation}
d_{ij}^{\mbox{\tiny BHAT}}= -\log{\frac{1}{K}\sum_{k=1}^K D_B(p_{ik},p_{jk})}.
\end{equation}


Other approaches could be used in order to define distances between the different transition matrices. For example, instead of using  $\mathbf{\tilde{A}}^n$ directly, an idea similar to diffusion distances \cite{szlam08} can be applied by using different powers of the transition matrices $\left(\mathbf{\tilde{A}}^n\right)^t$, where $t$ is a time index. This is equivalent to iterating the random walk defined by the transition matrices for $t$ time steps. The $j^{th}$ row of such an iterated transition matrix encodes the probabilities of transitioning from state $j$ to each other state in $t$ time steps. However, this introduces the extra parameter $t$, which must be set very carefully. For example, many transition matrices converge very quickly to the stationary distribution even for low $t$ (specially if the number of states is small). This can be a problem in cases where the stationary distributions for sequences in different clusters are the same. An example of such a scenario is presented in Section \ref{sec:results}.

Moreover, the SSD distance measure is very flexible. Measuring distances between sequences is highly subjective and application dependant. For example, in a certain scenario we may not be interested in the rest time for each state, but only in the transitions (similar to Dynamic Time Warping \cite{sakoe78}). To this end, a good alternative would be to obtain the transition matrices $\mathbf{\tilde{A}}^n$ for every sequence, but ignore the self transitions in the distance measurement. That can be easily done by setting all the self-transitions to 0 and then renormalizing the rows of the resulting transition matrices.

Once the distances between all the sequences are obtained, the actual clustering can be carried out using spectral clustering (or any other typical technique). We refer to this algorithm as state-space dynamics (SSD) clustering. It is summarized in Alg. \ref{alg:ssd}.
\begin{algorithm}[t]
\caption{SSD distance for clustering sequential data}
\label{alg:ssd}
\begin{algorithmic}
\STATE \emph{Inputs: }
\STATE Dataset $\mathcal{S}=\left\{\mathbf{S}_1,\ldots,\mathbf{S}_N\right\}$, $N$ sequences
\STATE $K$: Number of hidden states
\STATE
\STATE \emph{Algorithm:}
\STATE Step 1: Learning the global model (Baum Welch)
\STATE $\quad \theta=\arg\max_{\theta'}P(\mathbf{S}_1,\ldots,\mathbf{S}_N|\theta')$ 
\STATE
\STATE Step 2: Estimating $\mathbf{\tilde{A}}^n=\left\{\tilde{a}^n_{ij}\right\}$ (Forward/Backward)
\FORALL{$\mathbf{S}_n$}
\STATE $\alpha_k(t) = P(\mathbf{S}_n(1),\ldots,\mathbf{S}_n(t),q_t=k|\theta)$
\STATE $\beta_k(t) = P(\mathbf{S}_n(t+1),\ldots,\mathbf{S}_n(T_n),q_t=k|\theta)$
\STATE $\tilde{a}^n_{ij} \propto
\sum_{t=1}^{T_n} \alpha_i(t)a_{ij}p(\mathbf x_{t+1}|q_{t+1}=i)\beta_j(t+1)$
\ENDFOR
\STATE
\STATE{Step 3: Obtaining the distance matrix $\mathbf{D}=\left\{d_{ij}\right\}$}
\FORALL{$i,j$}
\STATE $p_{ik} \equiv  k^{th}$ row of $\mathbf{\tilde{A}}^i$
\STATE $d_{ij}= -\log{\frac{1}{K}\sum_{k,k'=1}^K \sqrt{p_{ik}(k')p_{jk}(k')}}$
\ENDFOR
\STATE
\STATE{Step 4: Clustering using $\mathbf{D}$ } 

\end{algorithmic}
\end{algorithm}

It is worth noting that our proposal does not include any special initialization of the large model representing the dataset, such as imposing a block-diagonal structure on the transition matrix to encourage the clustering \cite{smyth97}. We do not aim to obtain a single generative model of the complete dataset, but an adequate common representation that allows for a subsequent successful non-parametric clustering. 

An important free parameter of our method is the number of hidden states of the common model. It should be chosen accordingly to the richness and complexity of the dataset. In the worst case (that is to say, assuming that there is no state sharing amongst different groups), it should grow linearly with the number of groups. In this work, we have fixed this size a priori, but it could be estimated using well-known criteria such as BIC or AIC \cite{bishop06}. 
Remember that the forward-backward algorithm for HMMs is $O(K^2T)$, where $K$ is the number of states and $T$ the sequence length. This indicates that our proposal is specially suitable for datasets consisting of a large number of sequences coming from a small number of clusters, which is a usual case. In such a scenario, the number of hidden states required for a successful clustering is low, so the time penalty in the FW-BW algorithm will be more than compensated by the significant computational load reduction coming from the linear complexity in the number of sequences. If sequences coming from different clusters share some emission distributions, the improvements will be even more notorious, because the algorithm will exploit that sharing in a natural way.

Finally, our work can be seen as similar to \cite{ramoni02}. There, the authors propose a bayesian clustering method based on transition matrices of Markov chains. They assume that the sequences are discrete, so a Markov chain can be directly estimated via transition counts. Our proposal, on the other hand, uses Markov chains on the latent variables (states), what makes it far more general. Moreover, our focus is on defining a general model-based distance between sequences, so that the SSD distance can be directly coupled with a wide range of clustering algorithms depending on the task at hand.

\section{Experimental Results}
\label{sec:results}
In this section we present a thorough experimental comparison between SSD and state of the art algorithms using both synthetic and real-world data. Synthetic data include an ideal scenario where the sequences in the dataset are actually generated using HMMs, as well as a control chart clustering task. Real data experiments include different scenarios (character, gesture and speaker clustering) selected from the UCI-ML \cite{uci} and UCI-KDD \cite{ucikdd} repositories. The implementation of the compared algorithms is provided in the author's website\footnote{\url{http://www.tsc.uc3m.es/~dggarcia}}, except for PPK, which is available in Prof. Jebara's website\footnote{\url{http://www1.cs.columbia.edu/~jebara/code.html}}.

The compared methods for obtaining the distance matrix are: {\bf SSD},
state-space dynamics clustering with Bhattacharyya distance; {\bf PPK},
Probability Product Kernels \cite{jebara07}; {\bf KL}, KL-divergence
based distance \cite{garcia08}; {\bf BP}, BP metric \cite{panuccio02};
{\bf  YY}, Yin-Yang distance \cite{yin05} and {\bf SYM}, Symmetrized
distance \cite{smyth97}.

We denote the number of hidden states of the global model used by SSD as $K$, and the number of states per model of the methods that rely on training a HMM on each  single sequence as $K_m$.

Once a distance matrix is available, we perform the actual clustering
using the spectral algorithm described in \cite{ng02}. The different distance
matrices are turned into similarity matrices by means of a Gaussian
kernel whose width is automatically selected in each case attending to
the eigengap. Though more elaborated
methods such as [24] can be used to select the kernel width, in our experiments it is automatically selected in each case attending to the eigengap since the experimental results  are good enough.
We assume that the number of clusters is known a priori. If this
is not the case, automatic determination of the number of clusters can be carried out 
by methods such as those in \cite{sanguinetti05,zelnik04}.
The PPK method directly returns a similarity matrix,
that is first converted into a distance matrix by taking the negative
logarithm of each element. Then, it is fed into the clustering
algorithm with automatic kernel width selection. The final $k$-means
step of the spectral clustering algorithm is run 10 times, choosing as
the final partition the one with the minimum intra-cluster
distortion. The free parameter $T_{\mbox{ \tiny PPK}}$ of the PPK method is fixed to
10 following \cite{jebara07}.

The results shown in the sequel are averaged over a number of
iterations in order to account for the variability coming from the
EM-based training of the HMM. Performance is measured in the
form of clustering accuracy, understood as the percentage of correctly
classified samples under an optimal permutation of the cluster labels,
or its reciprocal, the clustering error.

\subsection{Synthetic data}
In this subsection we test the algorithms using two kinds of synthetically generated data: a mixture-of-HMMs (MoHMM) scenario as in \cite{smyth97,garcia08}, and a UCI-ML dataset representing control charts.
\subsubsection{Mixture of HMMs}
Each sequence in this dataset is generated by a mixture of two equiprobable HMMs $\theta_1$ and $\theta_2$. Each of these models has two hidden states, with an uniform initial distribution, and their corresponding transition matrices are
\begin{displaymath}
\mathbf{A}_1=\left( \begin{array}{cc}
0.6 & 0.4 \\
0.4 & 0.6
\end{array}\right)
\qquad
\mathbf{A}_2=\left( \begin{array}{cc}
0.4 & 0.6 \\
0.6 & 0.4
\end{array}\right).
\end{displaymath}
Emission probabilities are the same in both models, specifically $N(0,1)$ in the first state and $N(3,1)$ in the second. This is a deceptively simple scenario. Since both the emission probabilities and the equilibrium distributions are identical for both models, the only way to differentiate sequences generated by each of them is to attend to their dynamical characteristics. These, in turn, are very similar, making this a hard clustering task.
The length of each individual sequence is uniformly distributed in the
range $\left[0.6\mu_L,  1.4\mu_L\right]$, where $\mu_L$ is the mean length.

\begin{figure}[ht]
   \centering
   \includegraphics[width=9cm]{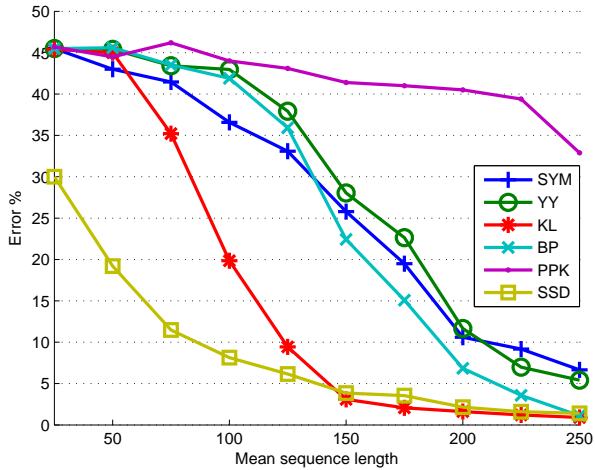}%
   \caption{Clustering error for the MoHMM case}
        \label{fig:synth}
\end{figure}

Figure \ref{fig:synth} shows the clustering error achieved by the compared algorithms in a dataset of $N=100$ sequences, averaged over 50 runs. All the algorithms use a correct model structure ($K_m=2$ hidden states per class) to fit each sequence. For SSD, this implies using 4 hidden states for the common model ($K=4$). As expected, when the sequences follow an HMM generative model and the representative model structure is chosen accordingly, SSD achieves impressive performance improvements for short sequence lengths. In contrast, algorithms that rely on training an HMM for each sequence suffer from poor model estimation when the mean sequence length is very low ($\leq 100$), which in turn produces bad clustering results. Our proposal overcomes this difficulty by using information from all the sequences in order to generate the common representative HMM. Consequently, the emission probabilities are estimated much more accurately and the distances obtained are more meaningful, leading to a correct clustering. Nonetheless, when the sequences are long ($\geq 200$) very accurate models can be obtained from each single sequence and the different methods tend to converge in performance. 

\subsubsection{Synthetic Control Chart}
\begin{figure}[t]
   \centering
   \includegraphics[width=9cm]{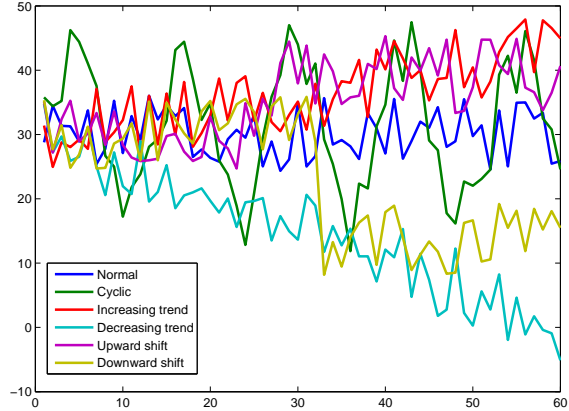}%
   \caption{Some samples from the Synthetic Control Chart dataset}
    \label{fig:ej_synth}
\end{figure}

This dataset contains unidimensional time series representing six different classes of
control charts: normal, cyclic, increasing trend, decreasing trend,
upward shift and downward shift. There are 100 instances of each
class, with a fixed length of 60 samples per instance. A sample of each class is plotted in Fig. \ref{fig:ej_synth}.

We carry out a multi-class clustering task on this dataset, partitioning the corpus
into 6 groups which we expect to correspond with the different classes
of control charts. As explained in Sec. \ref{sec:algorithm}, the size of the state-space for the HMM in SSD
clustering should be chosen accordingly to the number of classes, so
we employ a number of hidden states in the range 12-28. It
also allows us to show that our proposal is robust enough to produce
good performance in such an extense range. Results, averaged over 10 runs, are shown in Table
\ref{table:control}. It should be pointed out that the methods
SYM, YY,  KL,  BP, PPK could not handle the complete dataset of 100
instances per class in reasonable time, and had to be tested on a set of 30 sequences per
class. 

\begin{figure}[ht]
   \centering
   \includegraphics[width=9cm]{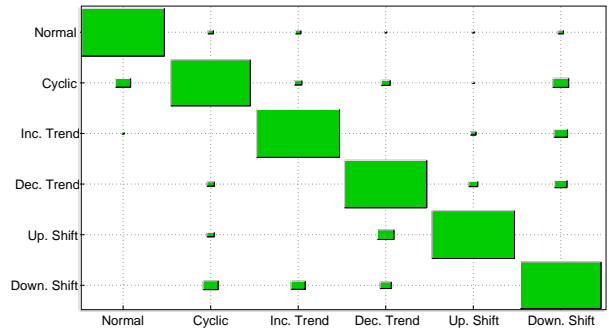}%
   \caption{Confusion matrix for SSD clustering with 30 sequences per class and 20 hidden states}
        \label{fig:confusion_matrix}
\end{figure}

SSD clearly outperforms the compared algorithms. It is also remarkable that these results confirm that our proposal benefits notably from larger datasets, as can be seen by comparing the performances for 30 and 100 sequences per class. This is due to the fact that, in contrast to previous proposals, the modeling employed by SSD clustering improves as the dataset size increases. The confusion matrix when $N=30$ and $K=20$ (averaged over the 10 runs) is shown in Fig. \ref{fig:confusion_matrix} in the form of a Hinton diagram.

\begin{sidewaystable}[tb]
\centering
\caption{Mean accuracy (standard deviation in brackets) in the Control Chart dataset. SYM, YY,  KL,  BP, PPK
had to be tested with 30 sequences per class due to computational reasons.}
    \label{table:control}
    \begin{tabular}{c|c c c c c}
        \hline
        \# hidden stat. & SYM & YY &  KL & BP & PPK \\
        \hline
        $K_m$=2 & 76.11\% \small{($\pm 0.1$)}& 74.11\% \small{($\pm 5.3$)}& 74.67\% \small{($\pm 5.3$)}& 78.33\% \small{($\pm 0.2$)}& 46.89\% \small{($\pm 3.3$)}\\
        $K_m$=3 &    74.00\% \small{($\pm 0.6$)}& 74.67\% \small{($\pm 0.5$)}& 78.67\% \small{($\pm 3.9$)}& 75.89\% \small{($\pm 3.6$)}& 55.22\%  \small{($\pm 0.6$)}\\
        $K_m$=4 &    74.78\%    \small{($\pm 0.6$)}& 76.44\% \small{($\pm 1.6$)}& 79.33\% \small{($\pm 3.1$)}& 76.00\% \small{($\pm 4.4$)}& 51.22\% \small{($\pm 4.8$)}\\
        $K_m$=5 & 77.00\% \small{($\pm 1.8$)}& 79.11\% \small{($\pm 3.2$)}& 77.44\% \small{($\pm 4.9$)}& 79.78\% \small{($\pm 1.6$)}& 41.78\% \small{($\pm 3.1$)}\\
        $K_m$=6 & 74.67\% \small{($\pm 0.3$)}& 76.89\% \small{($\pm 3.1$)}& 76.22\% \small{($\pm 2.9$)}& 74.44\% \small{($\pm 2.0$)}& 40.11\% \small{($\pm 3.5$)}\\
        \hline
    \end{tabular}
    \begin{tabular}{c|c c}
        \hline
        \# of hidden states& SSD (30 seq/class) & SSD (100 seq/class) \\
        \hline
        $K$=12 & 81.61\% \small{($\pm 6.0$)} & 91.57\% \small{($\pm 1.5$)} \\
        $K$=16 & 86.28\% \small{($\pm 6.3$)} & 92.71\% \small{($\pm 1.8$)} \\
        $K$=20 & 87.33\% \small{($\pm 4.1$)} & 93.23\% \small{($\pm 1.4$)} \\
        $K$=28 & \textbf{88.19}\% \small{($\pm 4.7$)} & \textbf{94.07\%} \small{($\pm 1.1$)} \\
    \end{tabular}
       
\end{sidewaystable}

\subsection{Real-world data clustering experiments}
We use the following  datasets from  the UCI ML and KDD archives:

\begin{description}
\item[Character Trajectories:] This dataset consists of trajectories captured by a digitizing tablet when writing 20 different characters. Each sample is a 3-dimensional vector containing the $x$ and $y$ coordinates as well as the pen tip force. The sequences are already differentiated and smoothed using a Gaussian kernel. We use 25 sequences per character, and carry out two-class clustering between all the possible combinations, giving a total of 190 experiments. The average length of the sequences in this dataset is around 170 samples.
\item[AUSLAN:] The Australian Sign Language dataset is comprised of 22-dimens\-ional time series representing different sign-language gestures. The gestures belong to a single signer, and were collected in different sessions over a period of nine weeks. There are 27 instances per gesture, with an average length of 57 samples. Following \cite{jebara07}, we perform 2-class clustering tasks using semantically related concepts. These concepts are assumed to be represented by similar gestures and thus provide a difficult scenario.
\item[Japanese Vowels:] To construct this dataset, nine male speakers uttered two Japanese vowels consecutively. The actual data is comprised of the 12-dimensional time-series of LPC cepstrum coefficients for each utterance, captured at a sampling rate of 10KHz using a sliding window of 25.6ms with a 6.4ms shift. The number of samples per sequence varies in the range 7-29 and there are 30 sequences per user. We use this dataset for two different tasks: speaker clustering and speaker segmentation.
\end{description}

\begin{table}[tb]
\small
\begin{minipage}{\textwidth}
\renewcommand{\arraystretch}{1.3}
\caption{Performance on the Character Trajectories (top), AUSLAN 
(middle) and Japanese Vowels (bottom, 9-class clustering task) datasets. The standard deviation of the results for the AUSLAN dataset is 0 in every case except for `SPEND' vs `COST' using YY distance, with a value of 0.8. The number of hidden states is $K=4$ for SSD and $K_m=2$ for the rest of methods (best case)}

\label{table:real}
\centering
\begin{tabular}{c|c c c c c||c|c} 
\hline
 \# hidden stat. & SYM & YY &  KL & BP & PPK &\# hidden stat. & SSD\\
\hline
\multirow{2}{*}{$K_m=2$}
& 96.10\%   & 97.02\% 	  & 96.42\%		  & 96.57\%     & 76.72\%      & \multirow{2}{*}{$K$=14} & 97.17\%\\
&\small{($\pm 0.4$)} & \small{($\pm 0.2$)} & \small{($\pm 0.1$)} & \small{($\pm 0.2$)} & \small{($\pm 0.8$)} &  &  \small{($\pm 0.3$)}\\

\multirow{2}{*}{$K_m=3$}
& 	96.30\% & 96.90\% & 96.45\% & 95.23\% & 67.66\% & \multirow{2}{*}{$K$=16} & 97.58\% \\
& \small{($\pm 0.1$)}& \small{($\pm 0.2$)} & \small{($\pm 0.0$)} & \small{($\pm 0.3$)} & \small{($\pm 0.8$)} & & \small{($\pm 0.2$)} \\

\multirow{2}{*}{$K_m=4$} &	95.31\%	&	95.53\% & 95.69\%& 83.92\% & 62.25\% & \multirow{2}{*}{$K$=20} & 98.23\%\\
& \small{($\pm 0.2$)} & \small{($\pm 0.0$)} &  \small{($\pm 0.1$)} & \small{($\pm 0.8$)} & \small{($\pm 0.2$)} & &  \small{($\pm 0.2$)} \\

\multirow{2}{*}{$K_m=5$} & 96.28\% & 96.40\% & 96.58\% & 84.70\% & 61.39\% & \multirow{2}{*}{$K$=22} & \textbf{98.35\%} \\
& \small{($\pm 0.3$)} & \small{($\pm 0.1$)} & \small{($\pm 0.1$)} & \small{($\pm 0.1$)} & \small{($\pm 0.2$)} & & \small{($\pm 0.2$)} \\
\hline
\end{tabular}

\begin{tabular}{c|c c c c c c} 
\hline
 SIGNS & SYM & YY &  KL & BP  & PPK &  SSD \\
\hline
`HOT' vs `COLD' & \textbf{100\%} & \textbf{100\%} & \textbf{100\%} & \textbf{100\%} & \textbf{100\%} & \textbf{100\%} \\
`EAT' vs `DRINK' &	51.85\%	&	92.59\%	&	92.59\% & 92.59\% & 93\% & \textbf{95.37}\%\\
`HAPPY' vs `SAD' &	59.26\% & 98.15\%	&	\textbf{100\%}	& 98.15\% & 87\% & \textbf{100\%} \\
`SPEND' vs `COST' & 54.07\% & 99.63\% & \textbf{100\%} & \textbf{100\%} & 80\% & \textbf{100\%} \\
`YES' vs `NO' & \textbf{60.36\%} & 55.56\% & 55.56\% & 55.56\% & 59\% & 56.66\% \\ 
\hline
\end{tabular}
\begin{tabular}{c|c c c c c||c|c} 
\hline
 \# hidden stat. & SYM & YY &  KL & BP & PPK &\# hidden stat. & SSD\\
\hline
\multirow{2}{*}{$K_m=2$}
& 66.67\%   & 85.11\% 	  & \textbf{90.15\%}		  & 85.30 \%     & 75.48\%      & \multirow{2}{*}{$K$=20} & 82.30\%\\
& \small{($\pm 2.8$)} & \small{($\pm 2.1$)} & \small{($\pm 1.7$)} & \small{($\pm 1.0$)} & \small{($\pm 3.7$)} &  &  \small{($\pm 3.7$)}\\

\multirow{2}{*}{$K_m=3$}
& 	67.18\% & 79.01\% & 81.14\% & 75.44\% & 82.59\% & \multirow{2}{*}{$K$=30} & 85.48\% \\
& \small{($\pm 3.0$)}& \small{($\pm 4.1$)} & \small{($\pm 4.0$)} & \small{($\pm 3.7$)} & \small{($\pm 5.1$)} & & \small{($\pm 4.5$)} \\

\multirow{2}{*}{$K_m=4$} 
&	67.96\%	&	83.41\% & 85.55\%& 78.55\% & 79.78\% & \multirow{2}{*}{$K$=40} & 87.93\%\\
& \small{($\pm 3.4$)} & \small{($\pm 6.2$)} &  \small{($\pm 5.4$)} & \small{($\pm 4.8$)} & \small{($\pm 3.7$)} & &  \small{($\pm 4.4$)} \\

\multirow{2}{*}{$K_m=5$} & 70.44\% & 82.81\% & 82.30\% & 79.77\% & 78.15\% & \multirow{2}{*}{$K$=50} & 86.37\% \\
& \small{($\pm 3.6$)} & \small{($\pm 5.2$)} & \small{($\pm 5.5$)} & \small{($\pm 4.4$)} & \small{($\pm 3.9$)} & & \small{($\pm 6.9$)} \\
\hline

\end{tabular}

\end{minipage}
\end{table}

Table \ref{table:real} shows the numerical results, averaged over 10 runs. In the Character Trajectories dataset, the individual sequences are fairly long and the classes are mostly well separated, so this is an easy task. Single sequences are informative enough to produce good representative models and, consequently, most methods achieve very low
error rates. Nonetheless, using the SSD distance outperforms the competitors.

For the AUSLAN dataset, following \cite{jebara07}, we used HMMs with $K_m=2$ hidden states for the methods that train a single model per sequence. The sequences were fed directly to the different algorithms without any preprocessing. We reproduce the results for the PPK method from \cite{jebara07}. The common model for SSD employs $K=4$ hidden states ($2K_m$), since the 2-way clustering tasks are fairly simple in this case. It is worth noting that the bad performance in the `Yes' vs `No' case is due to the fact that the algorithms try to cluster the sequences attending to the recording session instead of to the actual sign they represent. Our proposal produces great results in this dataset, surpassing the rest of the methods in every pairing except for the pathological `Yes' vs `No' case.

Finally, we carry out a 9-class speaker clustering task using the
Japanese Vowels dataset. The large number of classes and their
variability demands a large number of hidden states in the common HMM
of SSD. This, in turn, means a time penalty as the HMM training time is quadratic in the number of hidden states. Nonetheless, the performance obtained in this dataset by our proposal is very competitive in terms of clustering accuracy, only being surpassed by the KL method. It is also remarkable how the SSD-based clustering exhibits a very stable performance in a huge range of state-space cardinalities, what confirms our intuition that an accurate determination of that parameter is not crucial to the algorithm.

\subsubsection{Speaker Segmentation}
In order to briefly show another application of sequence clustering, we have reproduced the Japanese Vowels Dataset segmentation experiment from \cite{garcia09} using our proposed SSD distance. This scenario is constructed by concatenating all the individual sequences in the dataset to form a long sequence which we would like to divide into 9 segments (one per user). Only two methods, KL and SSD have been tested in this task because KL they were the best-performing methods in the clustering task. In order to carry out the sequence-clustering-based segmentation, we first extract subsequences using a non-overlapping window.  The length of these windows range from 10 to 20 samples. When using the KL distance, each subsequence is modeled using a 2-state HMM, which is the optimal value for the previously shown clustering task. A distance matrix is then obtained from these subsequences using the KL or SSD distances, and then spectral segmentation (SS) is carried out instead of spectral clustering \cite{garcia09}. This amounts to performing dynamic programming (DP) on the eigenvectors resulting from the decomposition of the normalized Laplacian matrix used for spectral clustering. In the ideal case, those eigenvectors are expected to be approximately piece-wise constant on each cluster, thus being a very adequate representation to run DP on in order to obtain a segmentation of the subsequences.

Table \ref{table:jap_seg} shows the performance obtained in this task. Segmentation error is measured as the number of incorrectly ``classified'' segments in the sequence, which can be seen as the fraction of the time that the segmentation algorithm is giving wrong results. As the results confirm, using the SSD distance is very adequate in this scenario because the number of subsequences is quite large and, at the same time, all of them are very short. This way, we exploit both the reduced time complexity in the number of sequences and the better estimation of the emission distributions. 

\begin{sidewaystable}[tb]
\caption{Segmentation error (mean and standard deviation) in the Japanese Vowels dataset (9 sources and segments) using KL and SSD distances}
\label{table:jap_seg}
\centering
\begin{tabular}{c|c | c c c c} 
\hline
& KL-SS & \multicolumn{4}{c}{SSD-SS} \\
Window length & $K_m$=2 & $K$=18 & $K$=24 & $K$=32 & $K$=40\\
\hline
W=10 & 	1.0\% ($\pm 0.22$) 	& 1.61\% ($\pm 0$)	& 1.0\%  ($\pm 0.29$)& 0.82\% ($\pm 0.35$)& 0.72\% ($\pm 0.33$) \\
W=15 &  3.5\% ($\pm 0$) 		& 2.39\% ($\pm 0$)  & 1.12\% ($\pm 0.39$)& 0.95\% ($\pm 0.29$)& 0.98\%  ($\pm 0.27$)\\
W=20 &	1.4\% ($\pm 0$) 		& 2.66\% ($\pm 0$)  & 1.26\% ($\pm 0.44$)& 1.03\% ($\pm 0.53$)& 0.75\%  ($\pm 0.45$)\\
\hline
\end{tabular}
\end{sidewaystable}

\section{Conclusions}
\label{sec:conclusions}
In this paper we have presented a new distance for model-based sequence clustering using state-space models. We learn a single model representing the whole dataset and then obtain distances between sequences attending to their dynamics in the common state-space that this model provides. It has been empirically shown that the proposed approach outperforms the previous semi-parametric methods, specially when the mean sequence length is short. Furthermore, the proposed method scales much better with the dataset size (linearly vs quadratically). As drawback of this method it should be mentioned that, as the number of classes grow, the common model may need a large number of hidden states to correctly represent the dataset (although the method is empirically shown not to be too sensitive to the accurate determination of the model size). In the case of hidden Markov models, the time complexity of the training procedure is quadratic in this number of states, so total running time can be high in these cases. Consequently, we find our proposal specially appealing for scenarios with a large number of sequences coming from a few different classes, which is a very usual case.

Promising lines for future work include the application of this methodology to other state-space models, both discrete and continuous, and to semi-supervised scenarios. We are also investigating alternative definitions of distance measures between transition matrices in order to take into account the potential redundancy of the state-space.











\end{document}